\documentclass[a4paper,twocolumn]{article}

\usepackage{jsaiac}
\usepackage{graphicx}
\usepackage{CJKutf8}
\usepackage{enumitem}
\usepackage{hyperref}       

\title{Multilevel Sentence Embeddings for Personality Prediction}

\address{paolo.tirotta@gmail.com,\,Akira.Yuasa@nttdata.com, moritams@nttdata-kansai.co.jp}

\author{%
Paolo Tirotta\first
\and
Akira Yuasa\first
\and
Masashi Morita\first
}

\affiliate{
\first{} NTT DATA
}

\begin{abstract}
Representing text into a multidimensional space can be done with sentence embedding models such as Sentence-BERT (SBERT). However, training these models when the data has a complex multilevel structure requires individually trained class-specific models, which increases time and computing costs.
We propose a two step approach which enables us to map sentences according to their hierarchical memberships and polarity. At first we teach the upper level sentence space through an AdaCos loss function and then finetune with a novel loss function mainly based on the cosine similarity of intra-level pairs.
We apply this method to three different datasets: two weakly supervised Big Five personality dataset obtained from English and Japanese Twitter data and the benchmark MNLI dataset. We show that our single model approach performs better than multiple class-specific classification models.
\end{abstract}

\def\BibTeX{{\rm B\kern-.05em{\sc i\kern-.025em b}\kern-.08em%
 T\kern-.1667em\lower.7ex\hbox{E}\kern-.125emX}}
\def\JBibTeX{\leavevmode\lower .6ex\hbox{J}\kern-0.15em\BibTeX}
\def\LaTeXe{\LaTeX\kern.15em2$_{\textstyle\varepsilon}$}

\begin{document}
\maketitle

\section{Introduction}
The rise of social networking services (SNS) in recent years has allowed market researchers and psychologists to access a huge user digital footprint to study the psychological characteristics of individuals. Knowledge of users' personality traits can be useful in everyday life, business, and healthcare, for example, in determining suitable social connections, targeting consumer markets, and even evaluating mental health \cite{vazire_carlson_2011}.

There is also a growing interest in the study of personality in the field of natural language processing, as studies \cite{mairesse_2007} have shown that written texts relate to the personality of the author. 
Nasukawa et al. \cite{nasukawa_2016} investigate linguistic features in Japanese texts that are highly relevant to the Big Five personality of the writer. Mori et al. \cite{mori_haruno_2020} investigated whether it is possible to estimate the personality of SNS posters by analyzing Twitter usage and post content, demonstrating the effectiveness of personality estimation using SNS information.

Large-scale Transformer-based language models (LLMs) such as BERT have emerged in Natural Language Processing (NLP) research and have outperformed previous methods on many tasks, so it can be expected that such models will also perform well in estimating psychology from social media. This paper focuses on the task of personality estimation, using the TIPI-J, the Japanese version of the Big Five personality scale, which measures personality as polarity values with negative, neutral, and positive orientations. These labels are assigned to short texts like tweets for each personality trait and serve as the ground truth. The task involves outputting polarity values ranging from -1 to 1 for each of the five personality axes.

Existing research has explored methods of encoding sentences with BERT and combining them with other models such as Stacked-NN \cite{leonardi_2020} or logistic regression and SVM \cite{arijanto_2021}. However, there has been no investigation into finetuning LLMs for the task of estimating polarity values in the five personality classes. In this task, the dataset has a two-level hierarchical structure with class labels for the five personalities and polarity labels for each class. Training a single model for multi-class learning and polarity learning is challenging using existing methods such as Sentence-BERT (SBERT). As a result, current methods require training separate models for each of the five personalities, leading to increased training time and resource costs.

We propose a two-step approach that allows mapping sentences according to the hierarchical memberships and polarities. First, the AdaCos loss function is used to learn the upper-level sentence space. Then the polarity is finetuned with a new loss function based on the cosine similarity of the sample pairs.
We also simultaneously investigate and evaluate the performance variation of different datasets by using this method on Japanese and English Tweet data for Big Five personality prediction and the English MNLI dataset. The results show that the proposed single-model approach yields better performance than multiple models trained with existing methods. 

The contribution of this study is to demonstrate the efficacy of a multi-stage deep metric learning approach with different loss functions in Japanese and English text classification tasks that involve hierarchically structured data, specifically the prediction of the Big Five personality traits. Our technique allows the development of a single-model approach that not only achieves improved performance, but also offers significant reductions in computational costs, eliminating the need for multiple models.

\section{Related Works}
\subsection{Sentence Embedding Models}
In this research, we employ Sentence Embedding Models to extract text features, as a means of converting the semantic meaning of a text into a numerical vector representation, referred to as an embedded representation. 
Previous methods, such as word2vec \cite{mikolov_2013}, Glove \cite{pennington2014glove}, and fastText \cite{bojanowski2017}, calculate this representation through the weighted average of pre-learned embedded representations of individual words. 
However, more recent approaches include bidirectional LSTMs such as InferSent \cite{conneau2017}, and Transformer models such as Universal Sentence Encoder (USE) \cite{cer2018} and SBERT. 
In a prior study \cite{reimers_2019}, a comparison of various models on the Semantic Textual Similarity task was conducted, with SBERT exhibiting the highest accuracy. Thus, we employ the SBERT method.

\subsection{Deep Metric Learning}
We train the SBERT model with various loss functions to optimize the embedded representations according to their respective classes. The loss functions that we compare can be classified into two main approaches: contrastive losses and angular losses.

\paragraph{Contrastive Losses. } Contrastive losses aim to learn the Euclidean distances between samples, such that samples with the same label are positioned closer together and those with different labels are positioned further apart in the feature space. A classic technique is Triplet Loss \cite{hoffer_2014}, which trains on three sets of data: a reference sample, a positive sample, and a negative sample.

\paragraph{Angular Losses. } 
Angular losses, on the other hand, focus on learning the angles in the feature space. The objective is to minimize the inner product between the embedded representations of a given data sample and its corresponding class. The Softmax Cross Entropy Loss is a widely used angular loss, and other state-of-the-art approaches, such as CosFace \cite{wang2018}, ArcFace \cite{deng2019}, and AdaCos \cite{zhang_2019}, enhance this concept by adding angular margins between classes. Generally, these approaches exhibit improved performance and stability compared to contrastive losses in supervised settings \cite{wang_2017}.

\paragraph{Limitations of Current Methods. } 
Both contrastive and angular losses are not well-suited to hierarchically structured data, as they do not explicitly model the hierarchical relationships of the data. As a result, the model can only try to guess the correct positioning of the sub-classes, without any explicit guidance on their polarity. One possible solution to this problem is to train multiple models, each responsible for a different level of the hierarchy. However, this approach can be costly and reduces the problem to a simple classification task.

\section{Datasets}
In this study, a weakly supervised personality dataset extracted from Twitter data and the benchmark MNLI dataset were used as data to train and verify the performance of the approaches introduced in the previous section. Details about dataset sizes can be seen in table \ref{tab:Table datasets}.

\subsection{Big Five Personality Dataset}
The Big Five personality traits, as defined in psychology, consist of five factors: agreeableness, neuroticism, conscientiousness, extraversion, and openness. 
In this study, tweets were collected based on each of these personality factors, with three polarities: positive, negative, and neutral.

\paragraph{Data Collection Procedure. }
At first, for positive and negative tweets, hashtags highly related to each personality trait were defined, and tweets were collected via Twitter searches using these hashtags as search queries. 

For neutral tweets we randomly collected tweets from the United States and Japan. These tweets were then processed through a pre-trained sentiment analysis model, and only tweets with a neutral sentiment were selected as neutral data.

\paragraph{Data Pre-processing. }
Initial pre-processing involved the removal of symbols, URLs, short sentences and other noises from the dataset. Additionally, many of the collected tweets contained advertisements and spam, resulting in hundreds of duplicate texts. To address this issue, the similarity between each text was calculated using TF-IDF, and tweets with similarity values above a fixed threshold were removed.

\paragraph{Test Data. }
In order to test the models on higher quality data, we prepared a test dataset composed of 100 sentences for both English and Japanese. Four human annotators scored these sentences on each of the five personality traits. The final labels are obtained as the average of the four individual scores.

\subsection{MNLI Dataset}
As part of the GLUE benchmark, the MNLI dataset \cite{williams_2017} has been used to evaluate various models and tasks. The train set consists of premise-hypothesis sentence pairs classified into five writing styles (telephone, government, slate, fiction, travel) and three label categories (entailment, neutral, contradiction). In this study, we utilize the writing styles as higher level categories and labels as polarity measures.
For testing, there are two different validation sets, the matched and mismatched sets. We employ the matched one as it contains genres that are consistent with the training set.

\begin{table}[h]
\centering
\caption{Size of the train and test sets for the three datasets: English and Japanese Personality and MNLI.}
\vspace{5pt}
\begin{tabular}{c|c|c|c}
\hline
Dataset           & EN Pers. & JP Pers. & MNLI  \\
\hline
Train Size         & 353k & 535k & 392k  \\
\hline
Test Size          & 500 & 500 & 9.8k \\
\hline
\end{tabular}
\label{tab:Table datasets}
\end{table}

\section{Proposed methodology}
To effectively represent hierarchical data in a space, an algorithm must be able to meet two key objectives. The first one is to map sentences having the same full hierarchy to a shared subspace, so that they have similar representations. Secondly, the algorithm must understand the relationships between different classes and sub-classes and reflect these differences as distances in the learned space. To this purpose, we combine the AdaCos loss and a novel loss based on the cosine similarity of sentence pairs. 

\begin{table*}[h!]
\centering
\caption{Results of four Sentence-BERT models trained with different loss functions: Triplet, Softmax, AdaCos and our approach. In addition, the results of five different classification BERT models are added as comparison. The evaluation metric is the Mean Absolute Error (MAE).} 
\vspace{5pt}
\resizebox{\textwidth}{!}{\begin{tabular}{c|c|c|c|c|c|c|c}
  \hline
 EN Pers. & Model & Agreebleness & Neuroticism & Conscientiousness & Extraversion & Openness & Average \\
 \hline
 single & Triplet & 0.427 & 0.480 & 0.382 & 0.406 & 0.404 & 0.420 \\
 single & Softmax & 0.464 & 0.550 & 0.373 & 0.471 & 0.430 & 0.458 \\
 single & AdaCos & 0.410 & 0.468 & 0.385 & 0.417 & 0.376 & 0.411 \\
 single & Ours & 0.396 & 0.411 & 0.409 & 0.349 & 0.342 & \textbf{0.381} \\
 5 models & BERT & 0.372 & 0.423 & 0.449 & 0.391 & 0.422 & 0.411 \\
  \hline
 JP Pers. & Model & Agreebleness & Neuroticism & Conscientiousness & Extraversion & Openness & Average \\
 \hline
 single & Triplet & 0.454 & 0.510 & 0.349 & 0.445 & 0.401 & 0.432 \\
 single & Softmax & 0.468 & 0.554 & 0.378 & 0.479 & 0.428 & 0.461 \\
 single & AdaCos & 0.437 & 0.491 & 0.311 & 0.400 & 0.347 & 0.397 \\
 single & Ours & 0.377 & 0.418 & 0.277 & 0.345 & 0.292 & \textbf{0.342} \\
 5 models & BERT & 0.377 & 0.504 & 0.479 & 0.454 & 0.406 & 0.456 \\
  \hline
 MNLI & Model & Slate & Government & Telephone & Travel & Fiction & Average \\
 \hline
 single & Triplet & 0.635 & 0.633 & 0.639 & 0.632 & 0.642 & 0.636 \\
 single & Softmax & 0.666 & 0.678 & 0.685 & 0.677 & 0.679 & 0.677 \\
 single & AdaCos & 0.589 & 0.562 & 0.587 & 0.576 & 0.565 & 0.576 \\
 single & Ours & 0.370 & 0.243 & 0.250 & 0.263 & 0.296 & \textbf{0.284} \\
 5 models & BERT & 0.384 & 0.268 & 0.316 & 0.312 & 0.326 & 0.321 \\
  \hline
\end{tabular}}
\label{tab:Table results}
\end{table*}

\subsection{AdaCos}
To accomplish the first objective, we use the AdaCos loss function to classify sentences based on their complete hierarchical structure. The use of AdaCos is driven by its ability to project sentences onto a hypersphere. This approach presents two main advantages compared to other methods.
\begin{enumerate}[itemsep=1pt,topsep=3pt]
    \item It allows us to place sub-classes with opposite polarities at polar opposites of the hypersphere.
    \item The space in the middle of the hypersphere is mostly empty and can be used to map neutral sub-classes, if present.
\end{enumerate}

\subsection{Pairwise Cosine Similarity Loss}
Following the AdaCos pre-training, in order to achieve the second objective, we formulate a cosine similarity loss between sentence pairs to teach the multilevel structure of the data and the distance between each sub-class. For the datasets under consideration, three main cases exist:
\begin{enumerate}[itemsep=1pt,topsep=3pt]
    \item Both sentences belong to the same class and have the same polarity: the assumed label is $1$.
    \item Both sentences belong to the same class, but have opposite polarity: the assumed label is $-1$.
    \item Both sentences belong to different hierarchies, or one of them has a neutral polarity: the assumed label is $0$.  
\end{enumerate}

In the third case, where the relationships between classes are not strongly supervised, we introduce a threshold $t$, which prevents the embeddings for neutral classes from amassing to the same region and allows some degree of freedom to the model's mappings. Losses with absolute value less than $t$ are considered as null. In our experiments, we find that a value of $t=0.3$ produces good results.

For a single batch, the overall loss is then calculated as the sum of each sentence pair loss divided by the number of comparisons in the batch.

\section{Results}
When evaluating a new sentence, we could calculate the Mean Absolute Error (MAE) between its true label and the class score, calculated as the cosine similarity between the new sentence embedding and the mean embedding of the target hierarchy.
However, this comparison method assumes that the higher-level class and polarity are known, which is not the case at test time. To address this issue, we estimate each class score as the average of the cosine similarity between the new sentence embedding and the positive and negative sub-classes mean embeddings.

The results in Table \ref{tab:Table results} show the comparison between our approach, the models trained with Triplet, Softmax and AdaCos losses, and five classification BERT models trained on each class. 
Compared to other methods, our proposed approach exhibits the lowest error across all datasets. 

Notably, for single-model approaches, we observe smaller differences for the annotated personality datasets, and larger ones for the MNLI dataset. This can be attributed to the fact that the scores of the annotated personality datasets are closer to neutrality, whereas the MNLI test data consists of hard labels. Furthermore, when checking the predictions for sentences with polarized scores in the personality datasets, only our approach and the multiple models' approach are able to correctly predict these sentences.
This is because our method effectively separates sub-classes based on their polarity, as we also demonstrate visually in the next section.

\subsection{Visualization}
\begin{figure}[h]
  \includegraphics[width=1\linewidth]{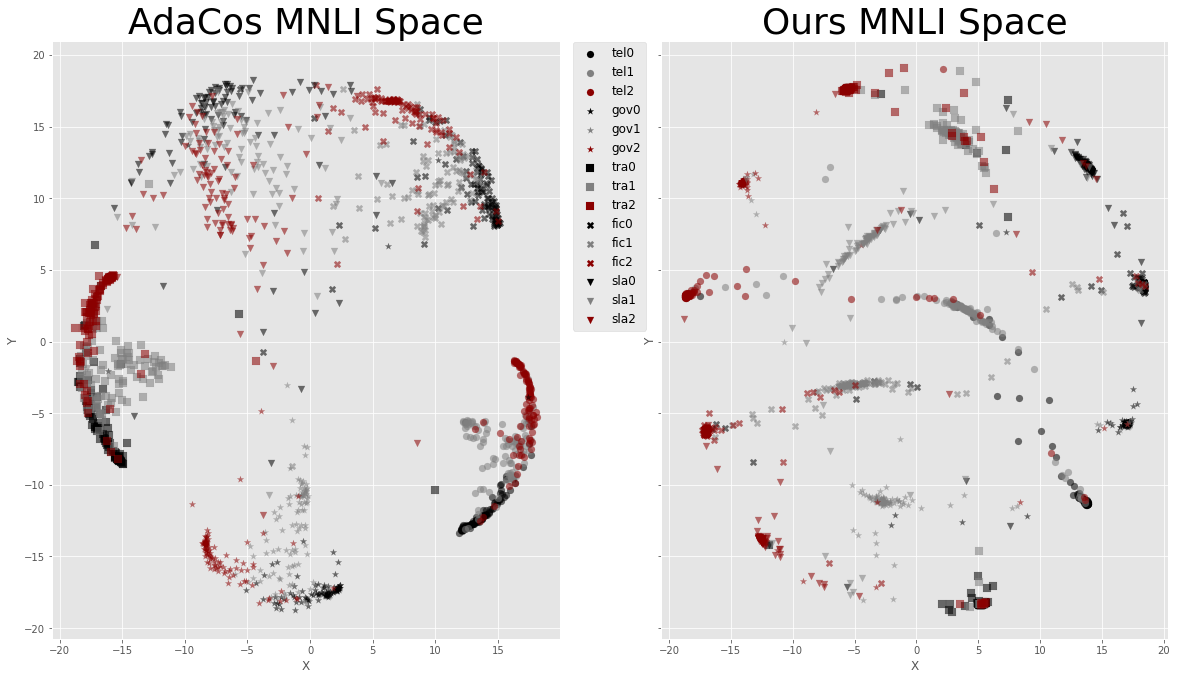}
  \caption{MDS space of the embeddings of a sample of MNLI data from models trained with AdaCos (left) and our (right) losses. Genres are classified by marker, while polarities are color-coded (positive: black, neutral: gray, negative: red).}\label{fig:adacos_adapair}
\end{figure}

To demonstrate the benefits of our method, we look at the effects of adding our Pairwise Cosine Similarity loss to the AdaCos trained model, and compare it with the models trained using Triplet and Softmax loss functions. 
We plot a sample of data points from each sub-class of the MNLI dataset using Multi Dimensional Scaling (MDS), as it represents the original multidimensional distances between points as faithfully as possible. These results can be found at our \href{https://github.com/Egojr/adapair}{Github repository} 
\footnote{Implementation and sample scripts for MNLI can be found at this repository: \href{https://github.com/Egojr/adapair}{https://github.com/Egojr/adapair}}.

On the left of Figure \ref{fig:adacos_adapair}, we can observe that the model trained with AdaCos loss shows low overlap between different upper-classes and that the center of the space is mostly empty. However, it is not suitable at representing the polarities of low-level classes as distances, as it groups them together depending on the higher-level class.

We address this issue through the Pairwise Cosine loss training, as we map sentences with neutral polarity to the middle of the space, and sentences with opposite polarity to opposite sides of the hypersphere. This process forms elongated shapes for each class depending on the polarity, with minimal overlap between all classes. 

\begin{figure}[h]
  \includegraphics[width=1\linewidth]{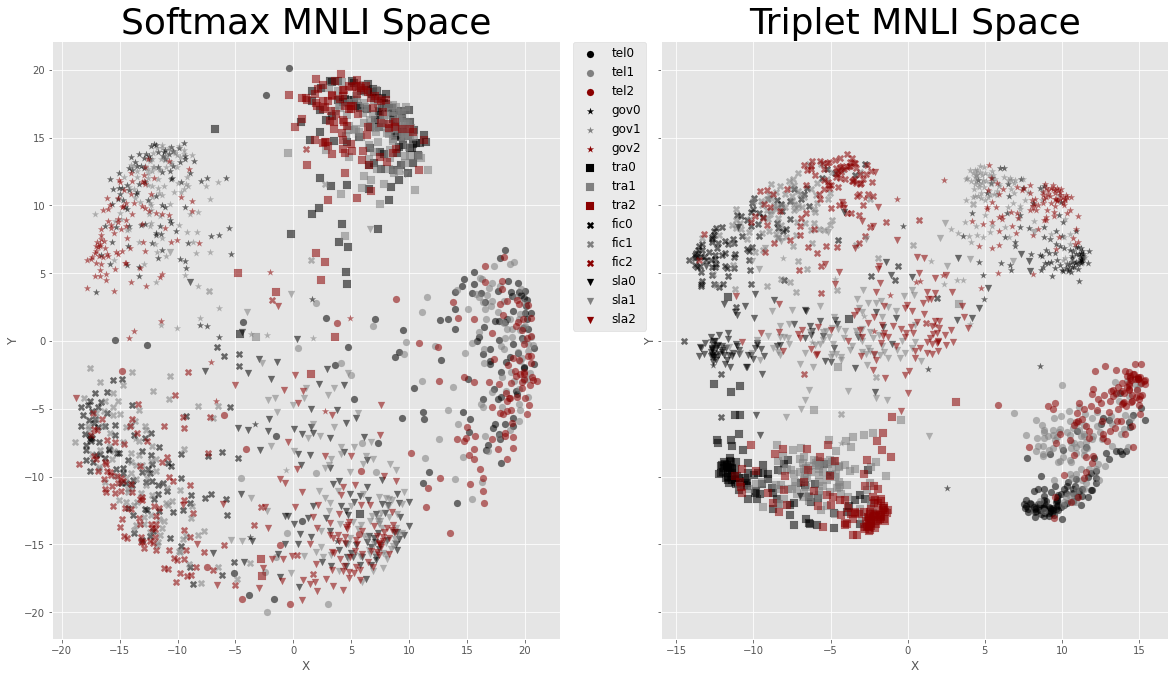}
  \caption{MDS space of the embeddings of a sample of MNLI data from models trained with Softmax (left) and Triplet (right) losses. Genres are classified by marker, while polarities are color-coded (positive: black, neutral: gray, negative: red).}\label{fig:soft, triplet}
\end{figure}

Models trained with Softmax and Triplet losses follow a similar pattern as the AdaCos model but they have higher overlap between classes as can be seen in Figure \ref{fig:soft, triplet}. Further experiments also highlighted this problem, as Pairwise Cosine models with pre-training from Triplet and Softmax trained models have higher error and overlap compared to finetuning the AdaCos pre-trained model.

\section{Conclusions}
Our findings show that current deep metric learning approaches are not suitable for training a single model to accurately map hierarchically structured data. Instead, a novel method combining AdaCos and a Pairwise Cosine Similarity loss was found to be more effective in learning the correlations between each class and their polarities, resulting in higher accuracy and reduced computational costs for inference than multiple class-specific classification models. 
Moreover, knowledge of the relationships between classes and sub-classes can be integrated in the loss function specification, by training with specified distances between specific classes. Modifying the threshold value can also lead to a more strict or lenient training of the model depending on the use case.

Applying this method to real-world data may enable the estimation of complex individual psychological characteristics, which could have implications for marketing and healthcare organizations. 

\begin{CJK}{UTF8}{min}
\bibliographystyle{unsrt}
\bibliography{references}

\begin{thebibliography}{10}

\bibitem{vazire_carlson_2011}
Simine Vazire and Erika~N. Carlson.
\newblock Others sometimes know us better than we know ourselves.
\newblock {\em Current Directions in Psychological Science}, 20(2):104–108,
  2011.

\bibitem{mairesse_2007}
Francois Mairesse, Marilyn Walker, Matthias Mehl, and Roger Moore.
\newblock Using linguistic cues for the automatic recognition of personality in
  conversation and text.
\newblock {\em J. Artif. Intell. Res. (JAIR)}, 30:457--500, 09 2007.

\bibitem{nasukawa_2016}
那須川哲哉, 上條浩一, 山本眞大, and 北村英哉.
\newblock
  日本語における筆者の性格推定のための言語的特徴の調査.
\newblock {\em 言語処理学会第 22 回年次大会発表論文集}, pages
  1181--1184, 2016.

\bibitem{mori_haruno_2020}
Kazuma Mori and Masahiko Haruno.
\newblock Differential ability of network and natural language information on
  social media to predict interpersonal and mental health traits.
\newblock {\em Journal of Personality}, 89(2):228–243, 2020.

\bibitem{leonardi_2020}
Simone Leonardi, Diego Monti, Giuseppe Rizzo, and Maurizio Morisio.
\newblock Multilingual transformer-based personality traits estimation.
\newblock {\em Information}, 11(4):179, Mar 2020.

\bibitem{arijanto_2021}
Joshua~Evan Arijanto, Steven Geraldy, Cyrena Tania, and Derwin Suhartono.
\newblock Personality prediction based on text analytics using bidirectional
  encoder representations from transformers from english twitter dataset.
\newblock {\em International Journal of Fuzzy Logic and Intelligent Systems},
  21(3):310--316, 2021.

\bibitem{mikolov_2013}
Tomas Mikolov, Kai Chen, Greg Corrado, and Jeffrey Dean.
\newblock Efficient estimation of word representations in vector space, 2013.

\bibitem{pennington2014glove}
Jeffrey Pennington, Richard Socher, and Christopher~D Manning.
\newblock Glove: Global vectors for word representation.
\newblock In {\em Proceedings of the 2014 conference on empirical methods in
  natural language processing (EMNLP)}, pages 1532--1543, 2014.

\bibitem{bojanowski2017}
Piotr Bojanowski, Edouard Grave, Armand Joulin, and Tomas Mikolov.
\newblock Enriching word vectors with subword information.
\newblock {\em Transactions of the association for computational linguistics},
  5:135--146, 2017.

\bibitem{conneau2017}
Alexis Conneau, Douwe Kiela, Holger Schwenk, Loic Barrault, and Antoine Bordes.
\newblock Supervised learning of universal sentence representations from
  natural language inference data, 2017.

\bibitem{cer2018}
Daniel Cer, Yinfei Yang, Sheng-yi Kong, Nan Hua, Nicole Limtiaco, Rhomni~St
  John, Noah Constant, Mario Guajardo-Cespedes, Steve Yuan, Chris Tar, et~al.
\newblock Universal sentence encoder.
\newblock {\em arXiv preprint arXiv:1803.11175}, 2018.

\bibitem{reimers_2019}
Nils Reimers and Iryna Gurevych.
\newblock Sentence-bert: Sentence embeddings using siamese bert-networks, 2019.

\bibitem{hoffer_2014}
Elad Hoffer and Nir Ailon.
\newblock Deep metric learning using triplet network, 2014.

\bibitem{wang2018}
Hao Wang, Yitong Wang, Zheng Zhou, Xing Ji, Dihong Gong, Jingchao Zhou, Zhifeng
  Li, and Wei Liu.
\newblock Cosface: Large margin cosine loss for deep face recognition.
\newblock In {\em Proceedings of the IEEE conference on computer vision and
  pattern recognition}, pages 5265--5274, 2018.

\bibitem{deng2019}
Jiankang Deng, Jia Guo, Niannan Xue, and Stefanos Zafeiriou.
\newblock Arcface: Additive angular margin loss for deep face recognition.
\newblock In {\em Proceedings of the IEEE/CVF conference on computer vision and
  pattern recognition}, pages 4690--4699, 2019.

\bibitem{zhang_2019}
Xiao Zhang, Rui Zhao, Yu~Qiao, Xiaogang Wang, and Hongsheng Li.
\newblock Adacos: Adaptively scaling cosine logits for effectively learning
  deep face representations, 2019.

\bibitem{wang_2017}
Jian Wang, Feng Zhou, Shilei Wen, Xiao Liu, and Yuanqing Lin.
\newblock Deep metric learning with angular loss, 2017.

\bibitem{williams_2017}
Adina Williams, Nikita Nangia, and Samuel~R. Bowman.
\newblock A broad-coverage challenge corpus for sentence understanding through
  inference, 2017.

\end{thebibliography}
\end{CJK}

\end{document}